\documentclass[letterpaper]{article} 
\usepackage[]{aaai23}  
\usepackage{times}  
\usepackage{helvet}  
\usepackage{courier}  
\usepackage[hyphens]{url}  
\usepackage{graphicx} 
\usepackage{natbib}  
\usepackage{caption} 
\usepackage{algorithm}
\usepackage{algorithmic}
\usepackage{newfloat}
\usepackage{listings}
\DeclareCaptionStyle{ruled}{labelfont=normalfont,labelsep=colon,strut=off} 
\floatstyle{ruled}
\newfloat{listing}{tb}{lst}{}
\floatname{listing}{Listing}
\usepackage{amsmath,mathtools,amsfonts,mathrsfs,amssymb,bm}
\usepackage{latexsym}
\usepackage{url}
\usepackage{amsthm}
\usepackage{nicefrac}

\newcommand{\beitemize}{\begin{list}{$\bullet$}{}} 
\newcommand{\enitemize}{\end{list}}

\newcommand{\beenumerate}{\hspace{} \begin{enumerate}} \newcommand{\enenumerate}{\end{enumerate}}

\newcommand{\belist}{\begin{list}{$\bullet$}{}} 
\newcommand{\enlist}{\end{list}}

\newcommand{\argmax}{\operatornamewithlimits{argmax}}
\newcommand{\argmin}{\operatornamewithlimits{argmin}}

\newcommand{\var}{\text{var}}

\newtheorem{theorem}{{\bf Theorem}}

\newtheorem{definition}{{\bf Definition}}

\newcommand{\rs}{r^*_j}

\newcommand{\acronym}{\text{SI}} 

\usepackage{caption}
\title{Using Simple Incentives to Improve Two-Sided Fairness in Ridesharing Systems}
\author{
Ashwin Kumar, Yevgeniy Vorobeychik, William Yeoh\\
}
\affiliations{
     Washington University in St. Louis
     \\
    \{ashwinkumar, yvorobeychik, wyeoh\}@wustl.edu
}

\begin{document}

\maketitle
\sloppy
\allowdisplaybreaks

\begin{abstract}
State-of-the-art order dispatching algorithms for ridesharing batch passenger requests and allocate them to a fleet of vehicles in a centralized manner,
optimizing over the estimated values of each passenger-vehicle matching using integer linear programming (ILP). Using good estimates of future values, 
such ILP-based approaches are able to significantly increase the service rates (percentage of requests served) for a fixed fleet of vehicles.
However, such approaches that focus solely on maximizing efficiency can lead to disparities for both drivers (e.g., income inequality) and passengers (e.g., inequality of service for different groups).
Existing approaches that consider fairness only do it for naive assignment policies, require extensive training, or look at only single-sided fairness.
We propose a simple incentive-based fairness scheme that can be implemented online as a part of this ILP formulation that allows us to improve fairness over a variety of fairness metrics. Deriving from a lens of variance minimization, we describe how these fairness incentives can be formulated for two distinct use cases for passenger groups and driver fairness. We show that under mild conditions, our approach can guarantee an improvement in the chosen metric for the worst-off individual. 
We also show empirically that our \emph{Simple Incentives} approach significantly outperforms prior art, despite requiring no retraining; indeed, it often leads to a large improvement over the state-of-the-art fairness-aware approach in \emph{both} overall service rate and fairness.
\end{abstract}

\section{Introduction}\label{sec:Introduction}

In a ridesharing system, multi-capacity vehicles allow passengers to share rides with others or be added onto existing trips, and a single central agent aggregates all information and dynamically matches passenger requests to available vehicles. On-demand ridesharing has been gaining traction over the past few years as a solution to the growing need for urban mobility and, as a consequence, there has been extensive work on developing approaches for optimizing the efficiency of such systems.
Recent approaches to this problem use
dynamic programming and deep reinforcement learning to learn value functions for matching pools of passenger requests to available drivers, and 
have led to significant improvements 
in service rate (percentage of passenger requests served) as well as total passenger delay~\cite{ride_alonso,ride_neurADP,ride_zac,ride_mfrl}.

Optimizing the efficiency of the overall system, however, can lead to both geographic disparity in the quality of service for the passengers, potentially exacerbating historical inequalities, and income disparity of the drivers.
While fairness in ridesharing has been a subject of some prior discussion~\cite{rf_tradeoff_nanda2020,rf_tradeoff2_xu2020}, the particular issue of \emph{geographic disparities} has received less attention. 
\citet{rf_neurADP_fair2020} consider balancing overall efficiency and \emph{either} geographic or income fairness, but this approach loses a great deal of overall efficiency, and requires complete retraining of the deep reinforcement learning model for any change in hyperparameters, such as the relative importance of fairness. 

Instead of proposing yet another algorithm-specific approach to address fairness in ridesharing, we seek to develop a general framework that can be applied orthogonally to a large class of existing ridesharing approach. Specifically, we leverage the two-stage approach that many existing ridesharing algorithms~\cite{ride_alonso,ride_neurADP,ride_zacbenders} rely on: (1)~Identifying or learning good value functions for matches, and (2)~Optimizing over those value functions to find a good match. 
While one could incorporate fairness in the value function learning stage, doing this would undermine our goal of generality since different ridesharing approaches employ different ways of identifying or learning their value functions. However, since all approaches employ similar ILP-based optimization approaches to find good matches, incorporating fairness in the second stage would result in the generality that we seek. 

Therefore, in this paper, we propose \emph{Simple Incentives}~(SI), a general framework for including fairness in ILP-based matching, using any off-the-shelf value function approximations.
The key ingredient in SI is a novel linear measure of relative disparity that can be associated with individuals based on their group membership (e.g,~passengers' origin and destination pair or drivers' relative income level). We make the following contributions in this paper:
(1)~We show how we can derive a general form for this measure from the lens of minimizing variance in a metric of interest, following which we derive instantiations of passenger- and driver-side fairness functions.
(2)~We theoretically demonstrate that our approaches, under mild assumptions, provably improve the service rate of passengers from the most historically disadvantaged region as well as improve the income of the driver with the least historical income.
(3)~We empirically demonstrate the generality of SI by showing that it significantly improves fairness for a variety of ridesharing algorithms with minimal compromises on service rate. 
Additionally, through extensive experiments, we show that the best SI variant achieves significantly greater fairness than the state-of-the-art fair ridesharing approach, while at the same time yielding overall system efficiency (measured by the service rate) that is nearly inline with a state-of-the-art approach that maximizes efficiency and ignores fairness. 
(4)~We show that our passenger- and driver-side approaches can be combined to improve two-sided fairness in such systems.
(5)~Finally, unlike the existing state of the art~\cite{rf_neurADP_fair2020}, our approach is completely online, allowing ridesharing operators or policy makers to tune the tradeoff between fairness and efficiency in real time during execution.

\section{Related Work}
\label{sec:related}  

\textbf{Order dispatching in ridesharing:} Rideshare-matching has been extensively studied, and researchers have introduced methods that improve the quality of the matches made in terms of increasing the number of requests matched~\cite{ride_zacbenders,ride_ma2015}, reducing the pickup and detour delays~\cite{ride_alonso,ride_huang2014}, and increasing drivers' earnings~\cite{rf_Balancing_eff_fairness_lesmana2019}. The complexity of ridesharing algorithms increases with the increase in vehicle capacity and fleet size. As the runtime of real-time algorithms need to be relatively small, most existing work has either considered assigning one request at a time (sequentially) to available drivers for high capacities~\cite{ride_ma2015,ride_tong2018unified} or assigning all active requests together in a batch for a small capacity~\cite{ride_yu2019integrated,ride_zheng2018order}. The sequential solution is faster to compute but the solution quality is typically poor~\cite{uberblog}. \citet{ride_alonso} proposes ILP optimization approaches for assigning all active requests together for high-capacity ridesharing. \citet{ride_neurADP} and \citet{ride_zacbenders} further improve these approaches by including information about anticipated future requests while matching current batch of requests to available drivers. 

\smallskip \noindent 
\textbf{Fairness in ridesharing:} Researchers have evaluated ridesharing fairness from many viewpoints. For passengers, there has been work on addressing lack of transparency~\cite{rf_FairVOpt_wolfson2017}, using game-theoretic approaches to fairness~\cite{rf_nash_foti2019}, and benefit sharing by ensuring non-increasing disutility~\cite{rf_passenger_cost_benefit_sharing_2016}.
Driver-side fairness has also been explored from the economic perspective, by using a max-min approach to fairness to balance efficiency and fairness~\cite{rf_Balancing_eff_fairness_lesmana2019}, and by looking at fairness over longer periods of time by equalizing driver income proportional to the number of hours spent on the platform~\cite{rf_2side_fair_suhr2019}.
Fairness isn't restricted to monetary benefits, however. Motivated by demographic and geographic fairness concerns, recent work formulates a matching problem with parameters to trade profit for fairness in terms of discriminatory cancellations, looking at factors like start/end locations, race, gender, or age of passengers~\cite{rf_tradeoff_nanda2020} 
and drivers~\cite{rf_tradeoff2_xu2020}.

A work that is closest to ours is by \citet{rf_neurADP_fair2020}, which looks at disparate treatment of passengers and income disparity amongst drivers.
While they also look at geographic zones to quantify fairness for passengers, their approach requires the training of a neural network based value function to include the fairness term in the objective, making it costly to change parameters for fairness. Our approach presents an online way to address this problem, without retraining existing value functions. 
Further, our approach offers better tradeoffs between efficiency and fairness as compared to the existing approach, and we show this in our empirical evaluation.

\section{Dynamic Matching in Ridesharing Settings}

A matching algorithm for ridesharing receives as inputs a continuous stream of batches of requests from passengers $\mathcal{R}$ and the current state of all the taxis $\mathcal{V}$ operating in a street network $\mathcal{G}$.
The street network $\mathcal{G} = \langle \mathcal{L}, \mathcal{E}, c(e)\rangle$ is a graph containing locations $\mathcal{L}$ connected by roads $\mathcal{E}$, with a cost function  $c:\mathcal{E}\rightarrow  \mathbb{R}^+$ that defines the cost $c(e)$ of each edge $e \in \mathcal{E}$ in the graph, which commonly corresponds to the time needed by a taxi to traverse $e$.
A request is a tuple $r = \langle q, d, t\rangle$ that contains the pickup location $q$, dropoff location $d$, and the request arrival time $t$.
The vehicle $i$ is associated with a state $v_i= \langle l_i, p_i, c_i, U_i\rangle$ that includes its location $l_i$, current path $p_i$, capacity $c_i$, and the set of requests $U_i$ it is currently serving.

State-of-the-art approaches to this dynamic matching problem take a discrete-time multi-agent 
perspective~\cite{ride_alonso,ride_neurADP}.
In this framework, a decision-maker is faced in each time step with a set of outstanding requests, along with the state of vehicles at that point in time.
The set of actions of this decision-maker is the set of all feasible matches of requests to vehicles.
Since any matching impacts the distribution of future requests (unmatched requests can spill over into the next time step) and vehicle states, the associated expected future value is captured by a value function.
This allows the decision-maker to make non-myopic decisions by considering expected rewards in addition to immediate rewards.
To manage problem complexity, a value function is associated with individual vehicles and assumed to be identical for all vehicles, conditional on vehicle state; furthermore, it is approximated by a \emph{Value Function Approximator} (VFA) -- most recently, by a deep neural network in the NeurADP framework~\cite{ride_neurADP}.

\begin{figure*}[t]
    \centering
    \includegraphics[width=0.65\linewidth]{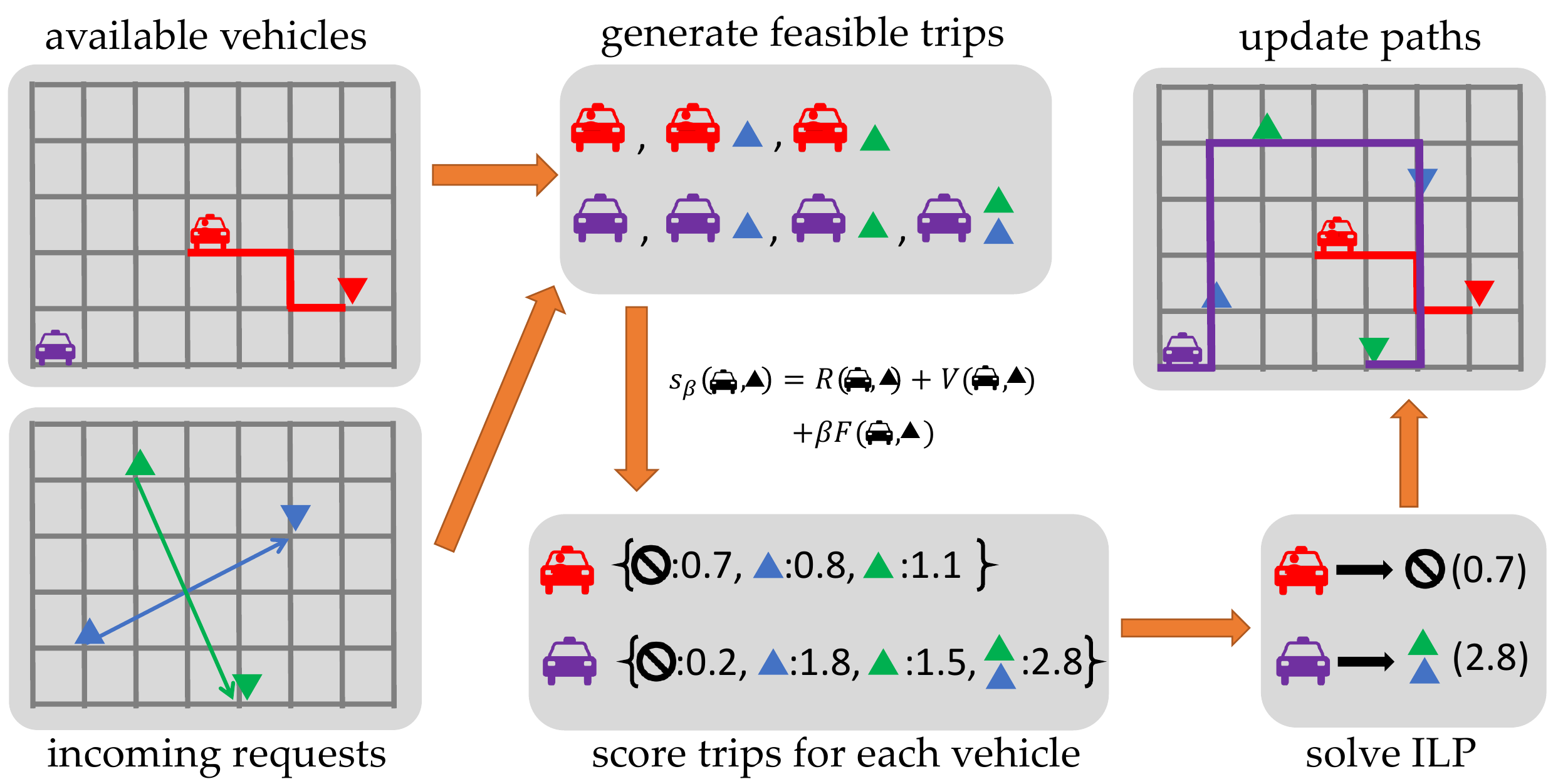}
    \caption{The rideshare-matching pipeline. Every time step, based on incoming requests and available vehicles, feasible trips for each vehicle are generated. These are assigned values using a score function, and an ILP is solved to optimize for total score.}
    \label{fig:pipeline}
\end{figure*}

A key workhorse in such approaches
is an integer linear program (ILP) for computing the optimal matching of requests to vehicles in a given time step, given a value function.
We now formalize the general form of this ILP, which is central to our proposed approach below.
For each vehicle $i \in \mathcal{V}$, we define an \emph{action} $a \subseteq \mathcal{R}$ as a subset of requests that could be matched to it.
Each action $a$ of vehicle $i$ is associated with a (discounted) value $V(i,a)$ obtained using VFA, and an immediate reward:
\begin{align}
\label{reward}
R(i,a) = \sum_{r \in a} R(i, r)
\end{align}
where $R(i, r)$ is an immediate reward to vehicle $i$ for servicing request $r$.
Let $A_i$ be the set of feasible actions of vehicle $i$ (i.e.,~all feasible subsets of requests that can be matched to $i$).
Then, for any feasible action $a \in A_i$ of vehicle $i$, we define the score of vehicle $i$ associated with action $a$ as:
\begin{align}
\label{score}
    s(i,a) = V(i,a) + R(i,a).
\end{align}

Let $\mathcal{A} = \{a_i\}_{i \in \mathcal{V}}$, $a_i \cap a_j = \emptyset$ for all $i \ne j$ be a matching of requests to vehicles.
We say that this matching is optimal if it maximizes the total score $\sum_i s(i,a_i)$.
We can compute such an optimal matching by solving the following ILP:

\begin{align}
    \max_{x_i(a) \in \{0,1\}} \sum_{i \in \mathcal{V}} \sum_{a\in A_i} x_i(a) & s(i,a) \quad \text{s.t.}  \label{eq:ILP} \\
    \sum_{a\in A_i} x_i(a) &= 1  &\forall i \in
    \mathcal{V} \label{eq:con1}\\
    \sum_{i \in \mathcal{V}} \sum_{a \in A_i| r \in a} x_i(a) &\le 1  &\forall r \in \mathcal{R}\label{eq:con2}
\end{align}
where $x_{i}(a)$ is an indicator variable associated with vehicle $i$ and its action $a \in A_i$. The constraints ensure that each vehicle is assigned exactly one action (Eq.~\ref{eq:con1}) and no request is assigned to more than one vehicle (Eq.~\ref{eq:con2}). In each vehicle's set of available actions, there is always the null action (i.e.,~accepting no new requests), so that there is always a solution. The final assignment $\mathcal{A}$ is a concatenation of all vehicle assignments. 
In existing approaches~\cite{ride_neurADP}, 
this ILP serves to maximize the total expected number of requests served (i.e.,~the \emph{service rate} when normalized by the total number of requests), and it can be readily generalized to maximize total expected profit.
Figure \ref{fig:pipeline} shows an illustration of the ILP matching process.

\section{\acronym: A General Fairness Framework}

By focusing on maximizing measures of efficiency such as service rate, traditional matching algorithms for ridesharing fail to account for the emergence of fairness issues for both the passengers and drivers. 
Additionally, since matching algorithms in ridesharing are often trained on real-world data, they can further exacerbate historical inequalities.
For example, from a passenger's perspective, it is conceivable that while maximizing efficiency, matching algorithms focus resources towards high-traffic areas (e.g.,~downtown locations) while reducing service in low-traffic areas (e.g.,~suburbs).
If such algorithms are deployed in practice, it may also lead to a feedback loop that continually reduces service to regions with low demand, which causes fewer passengers to make requests from such regions and so on. 
From a driver's perspective, the matching algorithms may prioritize drivers in high-traffic areas to continually service requests from those areas over drivers in low-traffic areas, leading to disparity in driver income. 
In this work, our goal is to tackle this kind of unfairness with minimal impact to the efficiency vis-a-vis the service rate.

We consider \emph{statistical parity}~\cite{fair_demoparity, fair_statparity_micro} as the notion of fairness in this work, which defines that the expected value of a given metric $z$ over a group $g$ is the same as $\bar{z} = \mathbb{E}_{g'\in G} \left [ z(g') \right ]$ the expected value of that same metric over all groups $g'\in G$:
\begin{align}
\label{eq1}
z(g) & = \bar{z}
\end{align}
More generally, we can write it as:
\begin{align}
    \left| \bar{z} - z(g) \right| &\le \epsilon  & \forall g\in G
\end{align}
where $\epsilon$ is a slack parameter.

As an example, let's say that we want parity in service rates for passenger groups defined by their origin-destination pairs and in normalized income for drivers. If a matching algorithm achieves statistical parity for both passengers and drivers, then the probability of a passenger receiving a ride is the same regardless of their origin and destination locations and the income of drivers are the same for all drivers. 

While the goal of achieving parity is noble, achieving it through the matching of a single time step is rarely possible, especially when there is a large disparity. 
Instead, it is often better to look at amortized parity over a longer period of time~\cite{rf_2side_fair_suhr2019}. 
To do this, we aim for a matching that ``moves closer'' towards parity with the goal of achieving parity in the near future. 
Towards that end, our framework uses variance $\var(\textbf{Z})$, where $\textbf{Z}=\{z(g), \forall g\in G\}$ is the set of metric values for all groups, as a proxy measure for fairness and, at each time step, it takes a gradient step in the solution space, moving in the direction that minimizes variance.

If we assume that the average of the metric over all groups is stable (i.e.,~$\frac{\partial}{\partial \mathcal{A}}\bar{z}\simeq 0$, a reasonable assumption if a long enough history is included), then we can find an assignment for a modified score function that accounts for the gradient of the variance with respect to the assignment $\mathcal{A}$:
\begin{align}
s'(i,a) 
 & = s(i,a) - \lambda \, \frac{\partial}{\partial \mathcal{A}} \var(\textbf{Z})  \\
 & = s(i,a) - \frac{1}{|\textbf{Z}|} \lambda \, \frac{\partial}{\partial \mathcal{A}} \sum_{z_j \in \textbf{Z}} (z_j - \bar{z})^2   \\
 & = s(i,a) + \frac{2}{|\textbf{Z}|} \lambda \, \sum_{z_j \in \textbf{Z}} (\bar{z} - z_j) \frac{\partial z_j}{\partial \mathcal{A}}  \label{eq:new-score}
\end{align}
where $\lambda$ is a hyperparameter. 
The general form above for the second term is our \emph{incentive} score: A constant (weight) multiplied by the disparity of group $j$, scaled by a derivative term. We show later that the sum can usually be simplified within the context of a given action $a$ and the derivative can be approximated for specific metrics, including our two passenger- and driver-side fairness metrics of interest.

The ``Simple Incentives'' idea is that, for each group involved in an action, provide them with an incentive (or penalty) proportional to how disadvantaged (or advantaged) their group has been historically.
Given the recent abundance of black-box algorithms, we find this simplicity helpful from a transparency perspective, making it easy to explain to any stakeholder how this score is calculated.

\smallskip \noindent \textbf{SI(+):}
While there is generally a consensus for applying incentives to help disadvantaged groups, it may be controversial to impose penalties for advantaged groups in some applications. With this in mind, we also present a modification to SI, where we clip the incentive term to be larger than $0$, resulting in the SI(+) variant, where the ``+'' indicates that we include only positive incentives.

In the following sections, we discuss how we specify the incentive score (see Eq.~\ref{eq:new-score}) for two use cases, one for passengers and one for drivers.

\subsection{SI for Passengers (SIP): Geographic Fairness}
\label{S:fair}

Our notion of passenger-side fairness is defined by the idea that the probability of receiving a ride should not depend on your origin or destination.
To do this, we divide the geographical area served by the fleet into a collection of areas $C$.
Recall that each request $r$ contains both the origin $o$ and destination $d$. Thus, we can map each passenger to one of $C \times C$ groups, uniquely identified by the origin-destination area pair based on $(o,d)$. Computing the service rate for each of these groups gives us the \emph{passenger}-side metric set $\textbf{Z}_p$, where $z_i\in \textbf{Z}_p$ denotes the \emph{historical} service rate for passenger group $i$.
For a given request $r$, let $g(r)$ denote the group to which $r$ belongs, and let $z(r)=z_{g(r)}$ (i.e.,~the historical service rate of the group $g(r) \in C \times C$).
These historical service rates are updated every time step after an assignment is made.
The goal is thus to achieve parity in the service rates for all geographic groups.

We now derive the fairness incentive for this metric, based on Eq.~\ref{eq:new-score}:
\begin{align}
s'(i,a)
& = s(i,a) + \frac{2}{|\textbf{Z}_p|} \lambda \, \sum_{z_j \in \textbf{Z}_p} (\bar{z} - z_j) \frac{\partial z_j}{\partial \mathcal{A}} \label{eq10} \\
& \simeq s(i,a) + \frac{2}{|\textbf{Z}_p|} \lambda \, \sum_{z_j \in \textbf{Z}_p} (\bar{z} - z_j) \frac{\partial z_j}{\partial a}  \label{eq11} \\
& = s(i,a) + \frac{2}{|\textbf{Z}_p|} \lambda \, \sum_{r \in a} (\bar{z} - z(r)) \frac{\partial z(r)}{\partial a}  \label{eq12} \\
& \simeq s(i,a) + \frac{2}{|\textbf{Z}_p|} \lambda \, \sum_{r \in a} (\bar{z} - z(r)) \, \lambda'   \label{eq13} \\
& = s(i,a) + \beta \, \sum_{r \in a} (\bar{z} - z(r)) \label{eq14} 
\end{align}
Notice that computing $\frac{\partial z_j}{\partial \mathcal{A}}$ is difficult due to circular dependencies: The matching $\mathcal{A}$ depends on the score function, which depends on the service rates, which in turn depend on the matching. As such, computing it precisely will require a global optimization procedure. Therefore, we approximate it with $\frac{\partial z_j}{\partial a}$ (from Eqs.~\ref{eq10} to~\ref{eq11}) based on the assumption that the change in the service rate of zone $z_j$ that is based on action $a$ is independent of the actions of other vehicles. 
Additionally, as serving a single request $r$ can only make small changes to the service rate $z(r)$ because service rates are aggregated over a reasonably long time period, we assume that the change is a (positive) constant $\lambda'$ for all requests (from Eqs.~\ref{eq12} to~\ref{eq13}).
We also introduce a \emph{fairness incentive} term $F_p(i,a)$ to represent the summation in the second term of Eq.~\ref{eq14} and use it in the new approximated score function~$s_\beta$: 
\begin{align}
F_p(i,a) &= \sum_{r \in a} (\bar{z} - z(r)) \label{eq15} \\
s_{\beta}(i,a) &= s(i,a) + \beta \, F_p(i,a) 
\end{align}
Here, $\beta$ is a hyperparameter that controls the relative importance of geographic fairness for passengers. In summary, for our SIP score function $s_\beta$, we include a fairness incentive term $F_p(i,a)$ that increases (or decreases) the score of actions proportional to how disadvantaged (or advantaged) their respective passenger groups have been according to our geographic fairness metric.

As described earlier, we also have the corresponding SI(+) version of this passenger-side incentive, which takes into account only requests from groups with below-average service rates:
\begin{align}
\label{eq:+r}
F_p(i,a) = \sum_{r \in a} \max\{\bar{z} - z(r),0\}
\end{align}
We call this variant of the framework SIP(+).

\subsection{SI for Drivers (SID): Income Fairness}

Our notion of driver-side fairness is defined by the idea that all drivers should earn similar incomes. This makes our driver-side fairness metric $\textbf{Z}_d$ the set of all $z_i$, where $z_i$ now denotes \emph{historical} driver incomes for each driver $i$. We scale the driver incomes by the largest driver income to restrict it to within $[0,1]$ and avoid scaling issues. The goal here is thus to achieve parity in the scaled income for all drivers. 

We now derive the fairness incentive for this metric, based on Eq.~\ref{eq:new-score}, where we model the income of a driver to be proportional to the value of the request they serve $R(i,r)$:
\begin{align}
s'(i,a)
& = s(i,a) + \frac{2}{|\textbf{Z}_d|} \lambda \, \sum_{z_j \in \textbf{Z}_d} (\bar{z} - z_j) \frac{\partial z_j}{\partial \mathcal{A}} \label{eq18} \\
& = s(i,a) + \frac{2}{|\textbf{Z}_d|} \lambda \, \sum_{z_j \in \textbf{Z}_d} (\bar{z} - z_j) \frac{\partial z_j}{\partial a} \label{eq19} \\
& = s(i,a) + \frac{2}{|\textbf{Z}_d|} \lambda \, (\bar{z} - z_i) \frac{\partial z_i}{\partial a} \label{eq20} \\
& = s(i,a) + \delta \, (\bar{z} - z_i) \, R(i,a) \label{eq21} 
\end{align}

\noindent Eqs.~\ref{eq18} and~\ref{eq19} are equivalent because, unlike the case in passenger-side fairness, computing $\frac{\partial z_j}{\partial \mathcal{A}}$ when the metric is driver income is straightforward as the income of a driver depends solely on their actions. Next, Eqs.~\ref{eq19} to~\ref{eq21} are equivalent because $\frac{\partial z_j}{\partial a} = R(i,a)$ when $j=i$ and is 0 otherwise.

Analogous to the case for passenger-side fairness, we also introduce a similar fairness incentive term $F_d(i,a)$ and use it in the new score function $s_{\delta}$:
\begin{align}
F_d(i,a) &=  (\bar{z} - z_i) \, R(i,a) \label{eq24} \\
s_{\delta}(i,a) &= s(i,a) + \delta \, F_d(i,a) 
\end{align}
Here, $\delta$ is a hyperparameter that controls the relative importance of income fairness for drivers.

We also have the variant that we call SID(+), where the fairness incentive is applied only to drivers with below-average income:
\begin{align}
\label{eq:+d}
F_d(i,a) = \sum_{r \in a} \max\{\bar{z} - z_i ,0\} \, R(i,r)
\end{align}

\section{Theoretical Properties}
While our approach directly works to minimize variance in the selected metric across groups, \emph{max-min} fairness is another popular notion of fairness, where we want to maximize the worst-off group. Towards that end, given a sufficient large weight ($\beta$ for passenger-side fairness and $\delta$ for driver-side fairness), our approaches provide guarantees for improving worse-off groups. We provide proof sketches here for the two metrics discussed in the previous section, with complete proofs our supplemental document.\footnote{
The code and supplementary material can be found at: \url{https://github.com/YODA-Lab/Simple-Incentives-For-Ridesharing} \label{ftnote:suppl_and_code}
} Both proofs assume that vehicles can only accept a single request at any time step (including the null action).
With this assumption, each action $a$ contains only one request $r$, allowing us to use them interchangeably.

Let $\mathcal{A}(w)$ be the matching solution produced by the ILP with fairness weight $w\in\{\beta,\delta\}$.
Further, let 
$z'_j(w)$ denote the updated metric value of group $j$ after an assignment $\mathcal{A}(w)$; and
$g_{\mathrm{min}} = \argmin_{j} z_j$ denote the group with the smallest metric value. We now describe the theoretical properties for each of our two fairness metrics below.

\subsection{Passenger-side Fairness Properties}

Let $\mathcal{R}_f=\{r\}_{g(r)=g_{\mathrm{min}}}$ denote the set of current requests corresponding to group $g_{\mathrm{min}}$. Any request $r \in \mathcal{R}_f$ would have the largest fairness incentive by definition (Eq.~\ref{eq15}).

In what follows, we state the theorem, claiming that as long as it is possible to improve the service rate of $z_{g_\mathrm{min}}$ relative to $\mathcal{A}(0)$ -- a condition that we formalize as \emph{passenger-min-unfairness} in the following definition -- we can do so for a sufficiently high $\beta$. 

\begin{definition}
\label{def:passenger-min-unfair}
A matching $\mathcal{A}$ is \emph{passenger-min-unfair} if there exists a request $r_f \in \mathcal{R}_f$ not served in $\mathcal{A}$, there is a vehicle $i \in \mathcal{V}$ such that $r_f \in A_i$,
but the request assigned to $i$ is $r_i \notin \mathcal{R}_f$.
\end{definition}
\begin{theorem}
If $\mathcal{A}(0)$ is passenger-min-unfair,  
then there exists $\beta > 0$ such that $z'_{g_\mathrm{min}}(\beta) > z'_{g_\mathrm{min}}(0)$.
\label{theorem:passenger-min}
\end{theorem}

\noindent \textbf{Proof Sketch:}
It can be shown that: 
\beitemize
\item [1.] The \emph{highest-preferred} request of vehicle $i$, $r_i^* = \argmax_{r \in A_i} s_{\beta}(i, r)$ will always be assigned to some vehicle in the optimal assignment. This follows from the nature of the ILP, which maximizes the total score. 

\item [2.] Any vehicle that prefers (see (1)) $r_f\in \mathcal{R}_f$ also prefers it for higher $\beta$ values. This can be proven using the fact that $r_f$ has the largest fairness incentive, and increasing $\beta$ only increases the contribution of the fairness incentive towards $s_{\beta}(i,a)$.

\item [3.] There is a threshold $\bar{\beta}$ value such that for higher $\beta$ values, the $\beta F_p(\cdot)$ term in $s_\beta(i,a)$ dominates the other factors (i.e.,~any action $a_1$ with $F_p(i,a_1)>F_p(i,a_2)$ will be preferred by vehicle $i$ over action $a_2$). This can be shown if we assume some bounds on $R(\cdot)$ and $V(\cdot)$.

\item [4.] Any request $r_f$ that was assigned in $\mathcal{A}(0)$ will also be assigned in $\mathcal{A}(\beta)$, where $\beta>\bar{\beta}\ge0$. This follows from a combination of the previous points. 
\enitemize
If $\mathcal{A}(0)$ is passenger-min-unfair, then we can conclude that there is some request $r \in \mathcal{R}_f$ that was not preferred in $\mathcal{A}(0)$ but will now be preferred (3) and assigned to some vehicle (1). Thus, $\mathcal{A}(\beta)$ assigns all requests in $\mathcal{R}_f$ that were assigned in $\mathcal{A}(0)$ and at least one more, thus increasing the service rate for $g_{\mathrm{min}}$ as compared to $\mathcal{A}(0)$. \hfill $\Box$

\subsection{Driver-side Fairness Properties}
For drivers, note that they would see a highest improvement in income after receiving an action with the highest immediate reward.
In what follows, we state the theorem, claiming that, for SID(+) with high enough $\delta$, any worse-off driver $j$ ($z_j<\bar{z}$) is guaranteed to get their highest preferred request $\rs$ as long as no other worse-off driver can serve it.
\begin{definition}[driver-min-unfair]
\label{def:driver-min-unfair}
A matching $\mathcal{A}$ is \emph{driver-min-unfair} if any worse-off driver $j$ is assigned a request $r\neq \rs$ in $\mathcal{A}$ and there exists no other worse-off driver $k$ that can serve $\rs$.
\end{definition}
\begin{theorem}
If $\mathcal{A}(0)$ is driver-min-unfair for any driver $j$,  
then there exists $\delta > 0$ such that $z'_j(\delta) > z'_j(0)$, when using SID(+).
\label{theorem:driver-min}
\end{theorem}

\noindent \textbf{Proof Sketch:} It can be shown that:
\beitemize
\item [1.] For a worse-off driver $j$, their highest-preferred request $\rs$ will have the highest score for some value of $\delta$, and will be assigned in $\mathcal{A}(\delta)$ to some driver.
\item [2.] Any better-off driver that can serve $\rs$ will not serve it for some large value of $\delta$. 
\enitemize
If $\mathcal{A}(0)$ is driver-min-unfair, then we can conclude that $\rs$ will be assigned to $j$ for some large value of $\delta$ because no other worse-off driver can serve it; no better-off driver can serve it~(2); and it must be served by some driver~(1). Thus, $z'_j(\delta) > z'_j(0)$ because $j$ is getting their highest-preferred request $\rs$ for some large $\delta > 0$, but not when $\delta = 0$. \hfill $\Box$

 \section{Experimental Evaluations}
\label{exp}

To comprehensively evaluate our SI framework, we ran three sets of experiments. First, to demonstrate the generality of the framework, we evaluated it in combination with various ridesharing algorithms, which use different value function approximations, from the literature. Second, to demonstrate the competitiveness of the framework, we evaluated it against a state-of-the-art ridesharing fairness approach. Finally, to demonstrate the flexibility of the framework, we evaluated it on two-sided fairness with both of our passenger- and driver-side fairness approaches combined. 

We evaluate the performance and fairness metrics after running the matching algorithms over a 24-hour period on the island of Manhattan using demand data from the NY Yellow Taxi dataset~\cite{yellowtaxi}. The locations in the road network correspond to street intersections, with edges as roads connecting them. 
We define areas with respect to a standard partition of Manhattan into neighborhoods, where passenger groups correspond to pairs of areas in which passengers are to be picked up and dropped off. Our efficiency objective is to maximize service rate, and thus, correspondingly, we set the value of each request to $1$.
For each hyperparameter $\beta$ and $\delta$, we performed a logarithmic search in the $[0.5, 20]$ range to capture a wide range of behaviors.
Consistent with literature~\cite{ride_neurADP,ride_zacbenders,rf_neurADP_fair2020}, we use a fleet size of 1000 vehicles, with a maximum request waiting time of 300 seconds. In all approaches, any request not assigned in the current assignment (one minute window) is dropped.
All experiments were run on a Ryzen 3700x CPU, RTX2080 Super GPU, and 32GB RAM.$^\text{\ref{ftnote:suppl_and_code}}$

We consider two standard measures of equity: The \emph{Gini} coefficient Gini(\textbf{Z}) and the minimum metric value min(\textbf{Z})\footnote{For drivers, we plot the minimum income (unscaled), to prevent bias because of scaling by the max.}
for a set $\textbf{Z}$ of service rates for passenger-side fairness or income for driver-side fairness. For ease of comprehension, we plot $F_{\mathrm{Gini}}(\textbf{Z}) = 1 - \text{Gini}(\textbf{Z})$ so that our goal is to maximize each of these metrics. Our efficiency measure is the overall service rate, defined as the fraction of all passenger requests served.

\begin{figure*}[t]
    \centering
    \includegraphics[width=\linewidth]{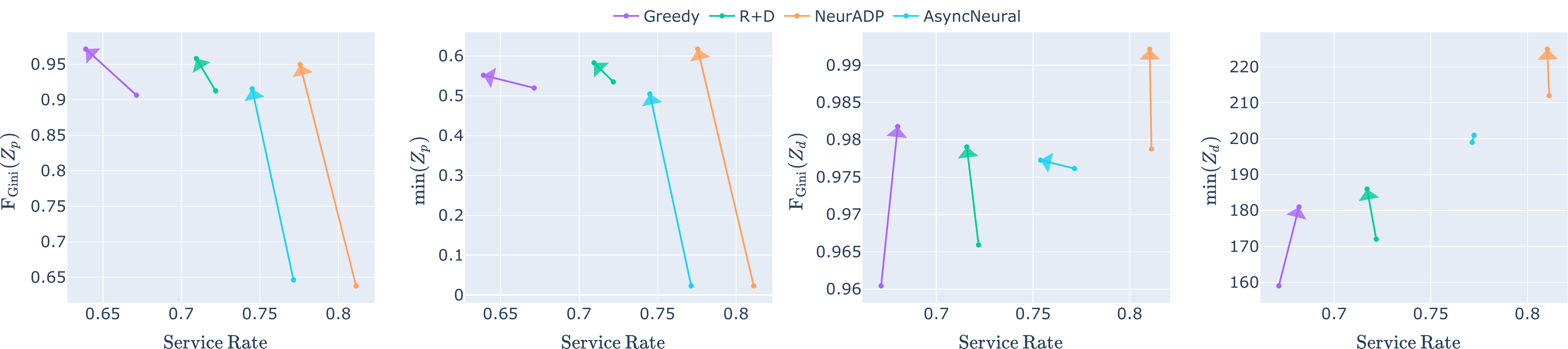}
    \hspace{2.5em} (a) \hspace{0.225\linewidth} (b) \hspace{0.225\linewidth} (c) \hspace{0.225\linewidth} (d)
    \caption{Change in efficiency and passenger-side fairness (with (a)~$F_{\mathrm{Gini}}$ and (b)~min) as well as driver-side fairness (with (c)~$F_{\mathrm{Gini}}$ and (d)~min) when Simple Incentives are used. The arrows show the best improvement for each algorithm metric while limiting SR to remain above 95\% of the initial value. The optimal point is to the top-right (high fairness and high service rate).}
    \label{fig:imp_vfs}
\end{figure*}

\begin{figure}[t]
    \centering \small
    \includegraphics[width=\linewidth]{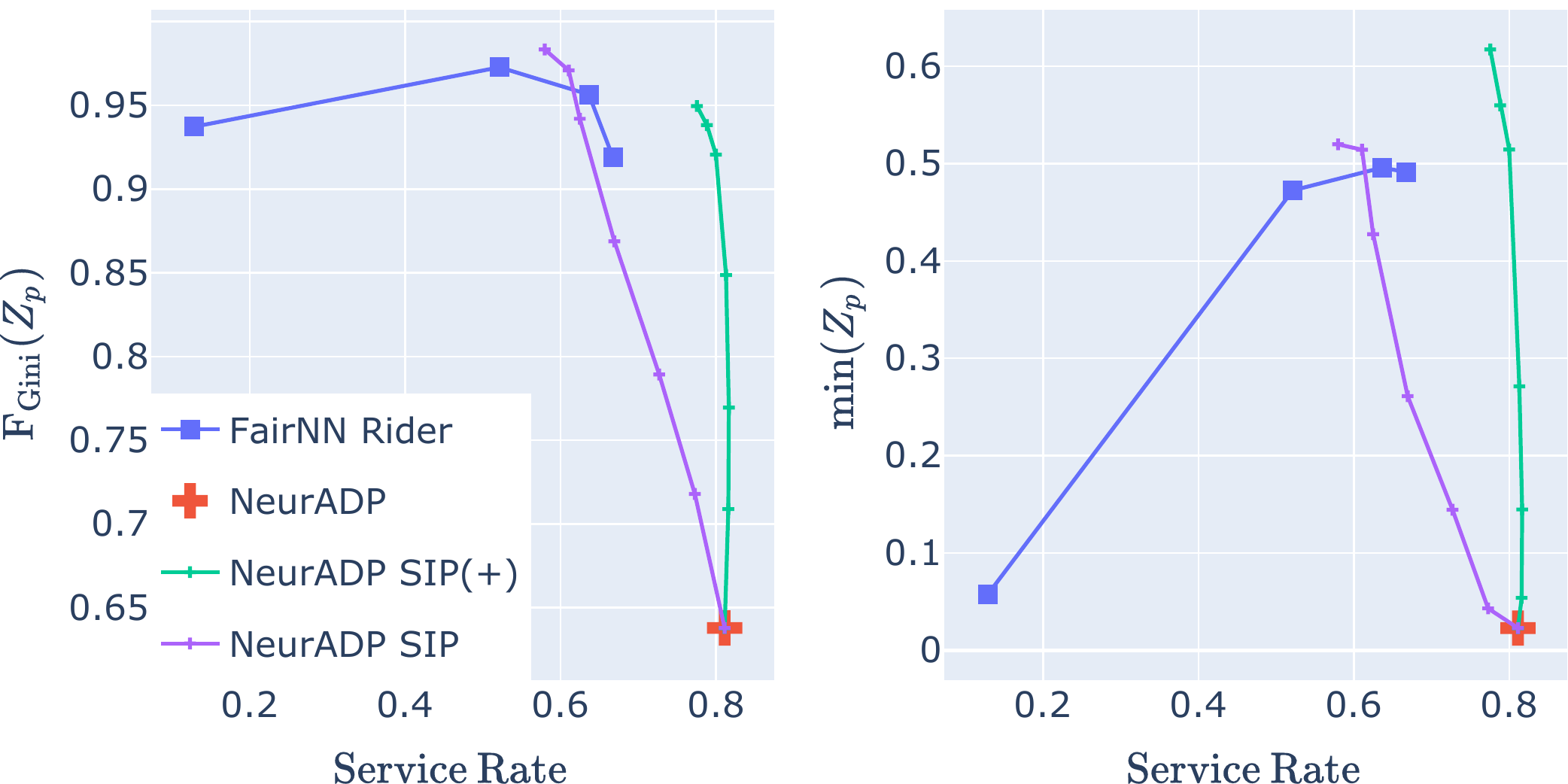}\\
    \hspace{2em} (a) \hspace{0.45\linewidth} (b)
    \caption{Comparison of SIP and SIP(+) against NeurADP and FairNN 
    (line plotted in the order of hyperparameter values) 
    for passenger-side fairness with (a)~$F_{\mathrm{Gini}}$ and (b)~min. The optimal point is to the top-right (high fairness and service rate).
    }
    \label{fig:paretofronts-passenger}
\end{figure}

\begin{figure}[t]
    \centering \small
    \includegraphics[width=\linewidth]{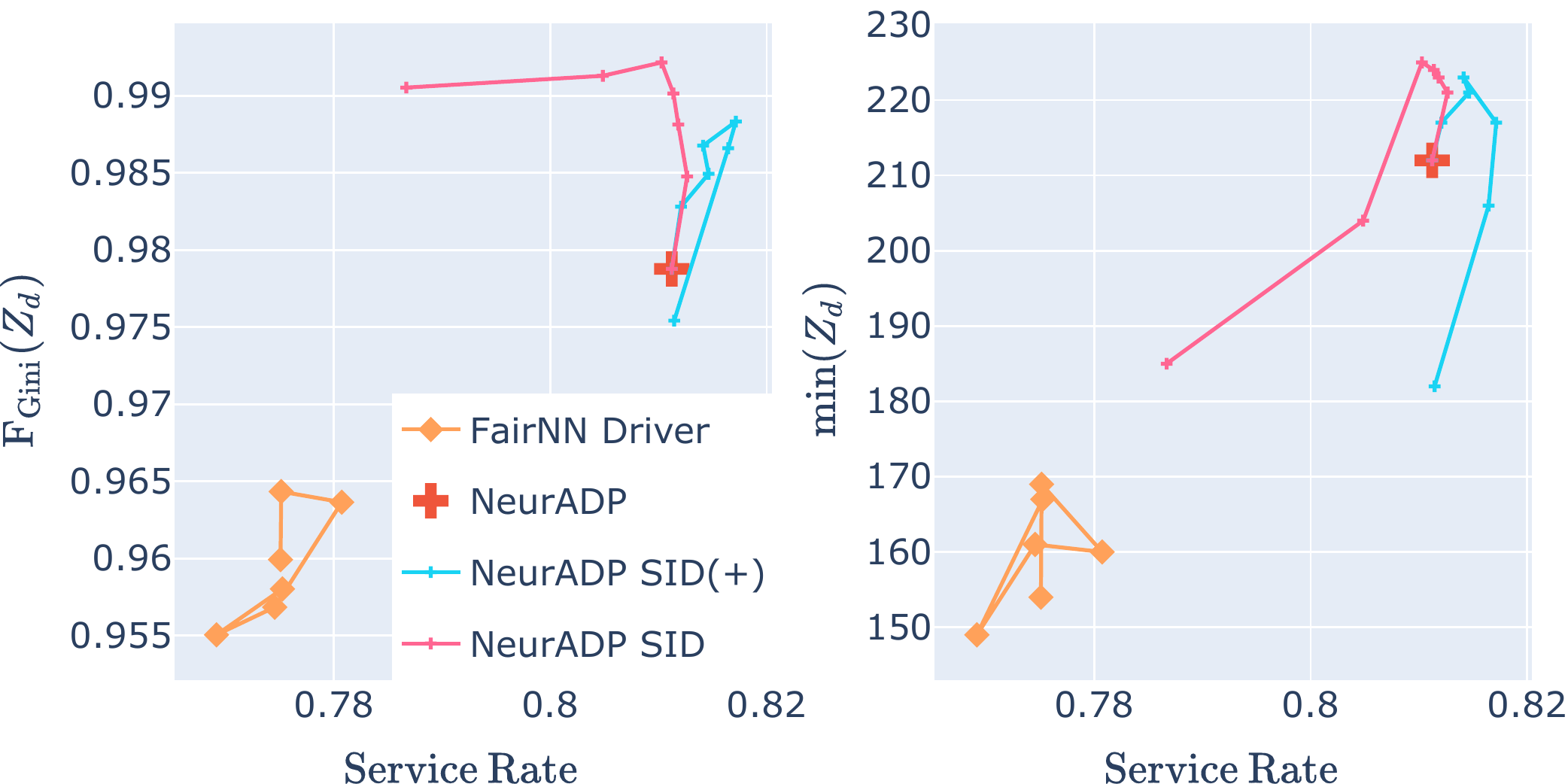}\\
    \hspace{2em} (a) \hspace{0.45\linewidth} (b)
    \caption{Comparison of SID and SID(+) against NeurADP and FairNN
    (line plotted in the order of hyperparameter values)  
    for driver-side fairness with (a)~$F_{\mathrm{Gini}}$ and (b)~min. The optimal point is to the top-right (high fairness and service rate).
    }
    \label{fig:paretofronts-driver}
\end{figure}

\begin{figure*}[t]
    \centering \small
    \includegraphics[width=\linewidth]{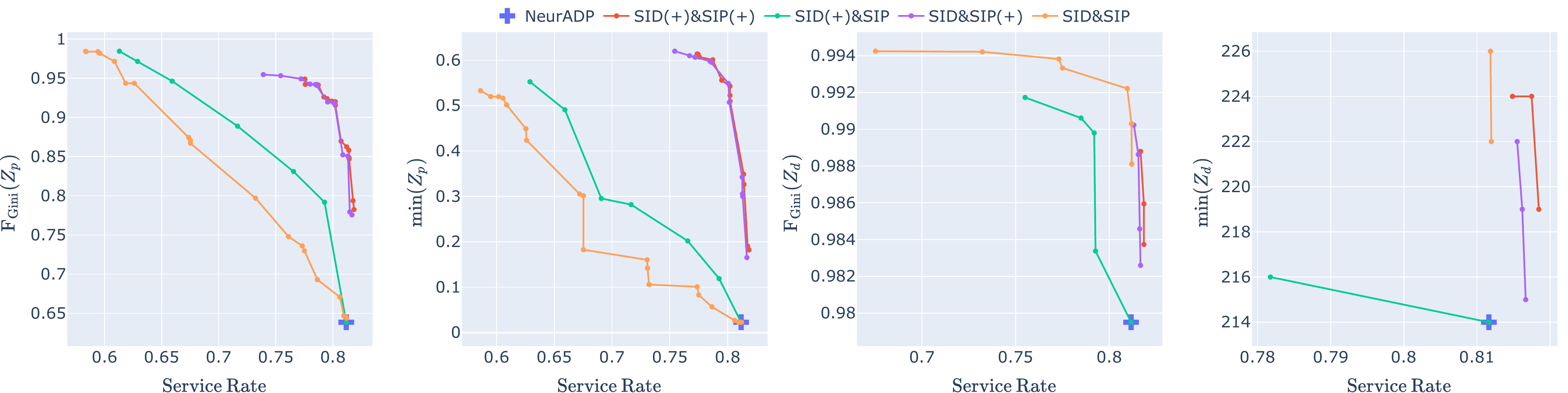}\\
    \hspace{2em} (a) \hspace{0.225\linewidth} (b) \hspace{0.225\linewidth} (c) \hspace{0.225\linewidth} (d)
    \caption{Comparison of Pareto frontiers of our combined variants on efficiency vs passenger-side fairness (with (a) $F_{\mathrm{Gini}}$ and (b) min) and driver-side fairness (with (c) $F_{\mathrm{Gini}}$ and (d) min). The optimal point is to the top-right (high fairness and service rate)
    }
    
    \label{fig:paretofronts-ablation}
\end{figure*}
\subsection{Generality: Evaluation on Benchmark Algorithms}

We tested the efficacy of our framework on a variety of natural and state-of-the-art ridesharing algorithms:
\beitemize 
\item \textbf{Greedy:} A baseline algorithm that considers only current rewards $R(i,a)$ as the score in the ILP optimization.
\item \textbf{R+D:} A pioneering approach proposed by \citet{ride_alonso} that estimates future values by using delays for passengers, as part of the score function in the ILP optimization. 
\item \textbf{NeurADP}: A state-of-the-art approach by \citet{ride_neurADP} that uses deep reinforcement learning to approximate the value function, as part of the score function in the ILP optimization. We use a pre-trained model trained with 1000 vehicles.
\item \textbf{AsyncNeural:} An asynchronous and distributed baseline algorithm that uses the approximated value function from NeurADP, but each vehicle greedily chooses its action and ties in vehicle order are broken randomly. (No ILP.)
\enitemize

Figure~\ref{fig:imp_vfs} show the impact of SI on our four ridesharing algorithms. The origin of each arrow corresponds to the performance of an algorithm without SI and the arrowhead corresponds to the best performance with SI while restricting the service rate to be above 95\% of the service rate without SI. Figures~\ref{fig:imp_vfs}(a) and~\ref{fig:imp_vfs}(b) show the results for passenger-side fairness for the $F_{\mathrm{Gini}}$ and min metrics, respectively; while Figures~\ref{fig:imp_vfs}(c) and~\ref{fig:imp_vfs}(d) show the results for driver-side fairness for those two metrics as well.

We observed that our methods can provide significant improvement in fairness at a marginal impact to service rates. Even with varying degrees of initial fairness, consistent improvement for both passengers and drivers can be seen. 
The one outlier is that SI failed to improve driver fairness for AsyncNeural. The reason is that, for each driver, their fairness incentive scaling term in $F_d(\cdot)$ for all actions are identical. Therefore, the incentive does not affect their preference ordering of requests to serve, and the outcome of the algorithm remains unchanged with and without the fairness incentive term. On the other hand, with the other centralized approaches, the ILP is able to arbitrate between different vehicles and prioritize low-income drivers.

\subsection{Competitiveness: Comparisons with FairNN}

As our fairness baseline, we use \emph{FairNN}, a recent fairness extension of NeurADP \cite{rf_neurADP_fair2020}. Like our SI framework, FairNN also considers geographic fairness for passengers \emph{or} income fairness for drivers, but unlike SI, it cannot consider both together. It does this by minimizing the variance in service rates or incomes, learning a neural network-based VFA to do so. As it follows a similar ILP formulation to solve the optimization problem, it is well suited for comparison with SI. 
It also has a hyperparameter $\lambda$ that controls the scale of the variance term in the objective, similar to $\beta$ or $\delta$ in our formulation.

We note below some key differences between our approach and FairNN:  
(1)~FairNN directly includes variance in the optimization, but as history size increases, the change in variance per action diminishes, reducing the impact of the fairness term if a constant weight is used. SI computes the scores based on historical inequalities rather than marginal contribution, which, combined with metric scaling, allows our approach to work well even with large histories.
(2)~FairNN applies the fairness uniformly across all actions, and we can improve \emph{both} overall efficiency and fairness by adding further flexibility to the objective (as shown below with 
$\acronym(
\textsc{+})$).
(3)~Possibly most important of all, this approach requires full retraining of VFA for even a small change in the tradeoff weight, or a change in any other problem parameters (such as the particular measure of fairness used), making it difficult to scale in practice. Our approach is completely online, and can be used with any pre-existing value function, and with any hyperparameter value.

We trained FairNN using their provided code, suggested parameter values, and suggested hyperparameter values $\lambda$ from $10^7-10^{10}$ for passengers and $\frac{1}{6} - \frac{6}{6}$ for drivers, each of which requires costly retraining.
For SI, we use the pre-trained NeurADP value function described in the previous section. NeurADP (without fairness) also acts as our efficiency baseline. 

\smallskip \noindent \textbf{Baseline Comparisons for Passenger-side Fairness:}
Figures~\ref{fig:paretofronts-passenger}(a) and~\ref{fig:paretofronts-passenger}(b) compare SIP and SIP(+), with different $\beta$ values, against both baselines.
FairNN outperforms SIP when the service rate is small ($\sim$0.65), but SIP(+) significantly outperforms FairNN by achieving similar fairness with much higher service rates. Consistent with results in Figure~\ref{fig:imp_vfs}, SIP and SIP(+) improves the fairness of NeurADP as $\beta$ increases.

\smallskip \noindent \textbf{Baseline Comparisons for Driver-side Fairness:} 
Figures~\ref{fig:paretofronts-driver}(a) and~\ref{fig:paretofronts-driver}(b) compare SID and SID(+), with different $\delta$ values, against NeurADP. 
FairNN for drivers had poor efficiency compared to NeurADP, which is consistent with observations by the authors~\cite{rf_neurADP_fair2020}. 
Performing better than FairNN, SID and SID(+) improve the fairness of NeurADP as $\delta$ increases. Analogous to the passenger-side variants, SID(+) outperforms SID. 

\smallskip \noindent \textbf{Relative Performance of our Variants:} 
For both drivers and passengers, we observed the SI(+) variants provided better tradeoffs between fairness and efficiency. The reason is that these variants improve fairness without significantly sacrificing efficiency since high-efficiency groups are not penalized. The base variants achieved much better fairness in the extreme, albeit at a higher cost to service rate. The reason is that if extreme fairness is required, then there may be no choice but to penalize high-efficiency groups.

\subsection{Ablation Experiments: Two-Sided Fairness}
Finally, we evaluate SI with NeurADP's value function on two-sided fairness with both passenger- and driver-side fairness approaches combined. Specifically, we ran ablation experiments, running all combinations of $\beta$ and $\delta$ values in a logarithmic grid search for the score function: 
\begin{align}
\!\!\!\!\!    s_{\beta,\delta}(i,a) = s(i,a) +\beta F_p(i,a) + \delta F_d(i,a)
\end{align}

Figure~\ref{fig:paretofronts-ablation} shows the Pareto frontier for each passenger- and driver-side fairness combination. In general, we observe the trend that using SIP leads to much lower service rates as compared to SIP(+). SID and SID(+) with SIP(+) are almost equivalent, Pareto dominating the other approaches, with the exception of extremely high fairness regions. SIP and SID combined resulted in the fairest algorithm, albeit at a high cost to service rate.

Typically, one would expect a tradeoff between fairness and service rate. However, out of 175 combinations of $\beta$ and $\delta$ hyperparameter values as well as SIP/SIP(+) and SID/SID(+) pairs, 15 combinations ($\sim$8.5\%) outperformed NeurADP on \emph{all} five metrics (four fairness metrics and service rate). A commonality across all 15 combinations is that they all have small hyperparameter values with $\beta \leq 2$ and $\delta \leq 2$.  
Further, 41 out of 175 combinations ($\sim$23.5\%) outperformed NeurADP on four out of five metrics, with $\beta \leq 2$ and $\delta \leq 10$.
This observation reinforces the idea that \emph{fairness does not have to be a trade off} and, in many cases, improving fairness and efficiency can go hand-in-hand.

\section{Discussion and Conclusions}

As the demand for cutting-edge algorithms for urban mobility increases, their effects on the underlying fairness of these systems need to be studied, and measures taken to ensure that the algorithms do not inherit implicit biases that result from pure optimization.
In this work, we focused on the issue of fairness in ridesharing systems, specifically geographic fairness for passengers and income fairness for drivers. We proposed Simple Incentives (SI), a general fairness framework that can be adapted to a class of existing ridesharing algorithms that use an integer linear program to find matches. At a high level, SI applies fairness incentives to groups of passenger requests and drivers based on their service rate and income disparity. Under mild assumptions, SI provably improves the service rate of passengers from the group with the worst service rate as well as the income of driver with the lowest income. Our experimental results demonstrated its generality by showing that it improves the fairness for several ridesharing benchmarks and its competitiveness by outperforming existing state of the art in terms of both efficiency (measured through overall service rates) and fairness (measured through the Gini coefficient and the minimum service rates and driver income) despite requiring no retraining. Our experiments showed that it is better to apply fairness incentives only to requests from passenger groups with below-average service rates and to drivers with below-average income, as opposed to a blanket approach that applies fairness incentives to all passenger requests and drivers. Finally, interestingly, our results showed that it is possible to improve both efficiency and fairness in some cases. 

\smallskip \noindent \textbf{Limitations:} 
We do not prescribe ``best'' hyperparameters as that decision is subjective and will depend on the use-case. Instead, we provide an easy way to tune the importance of fairness using the hyperparameters. We selected geographic fairness because of the lack of other protected demographic information of passengers in public datasets. However, our formulation is general enough to allow any group-based division of passengers (e.g.,~by race and/or gender). Our income fairness does not take into account the number of hours worked by drivers as we assume that all drivers work all 24 hours, consistent with the literature.
Finally, SI is a myopic approach to fairness. We may expect to see a non-myopic approach perform better when system dynamics allow for making suboptimal actions in the present that could lead to better fairness returns in the future. However, we expect these situations to be infrequent in practice.

\section*{Ethics Statement}

This work has the potential to affect livelihoods of taxi drivers in real ridesharing systems. We do not intend this work to be directly translated to application without sufficient testing to ensure that the effects on drivers' and other stakeholders' livelihood are not adverse.

\section*{Acknowledgements}
This research is partially supported by the National Science Foundation under awards 
1812619, 
1905558, 
1939677, 
2020289, 
2118201,
and 
2214141; 
by ARO grant W911NF1810208; and by Amazon. The views and conclusions contained in this document are those of the authors and should not be interpreted as representing the official policies, either expressed or implied, of the sponsoring organizations, agencies, or the United States government.

\bibliography{main_CR}

\end{document}